\documentclass{article}

\usepackage{microtype}
\usepackage{graphicx}
\usepackage{subcaption}
\usepackage{multirow}
\usepackage{booktabs} 
\usepackage{tabularx} 
\usepackage[table]{xcolor}
\usepackage{hyperref}

\usepackage[preprint]{icml2026}

\usepackage{amsmath}
\usepackage{amssymb}
\usepackage{mathtools}
\usepackage{amsthm}

\usepackage[capitalize,noabbrev]{cleveref}

\theoremstyle{plain}

\theoremstyle{definition}

\theoremstyle{remark}

\usepackage[textsize=tiny]{todonotes}

\icmltitlerunning{EVL-ECG: Efficient ECG Interpretation With Multi-Aspect Heterogeneous Knowledge Distillation}

\begin{document}

\twocolumn[
  \icmltitle{EVL-ECG: Efficient ECG Interpretation With Multi-Aspect Heterogeneous Knowledge Distillation}



  \icmlsetsymbol{equal}{*}

  \begin{icmlauthorlist}
    \icmlauthor{Nguyen Hong Dang}{hust,vinshc}
    \icmlauthor{Nhi Ngoc-Yen Nguyen}{vinshc}
    \icmlauthor{Huy-Hieu Pham}{vinshc,vineng,cihs}
  \end{icmlauthorlist}

    \icmlaffiliation{hust}{Hanoi University of Science and Technology, Hanoi, Vietnam}
    \icmlaffiliation{vineng}{College of Engineering and Computer Science, VinUniversity, Hanoi, Vietnam}
    \icmlaffiliation{vinshc}{VinUni-Illinois Smart Health Center, VinUniversity, Hanoi, Vietnam}
    \icmlaffiliation{cihs}{Center for Innovations in Health Sciences, VinUniversity, Hanoi, Vietnam}
    
    \icmlcorrespondingauthor{Huy-Hieu Pham}{hieu.ph@vinuni.edu.vn}

  \icmlkeywords{ECG Interpretation, Vision-Language Models, Knowledge Distillation}

  \vskip 0.3in
]



\printAffiliationsAndNotice{}  

\begin{abstract}
High-fidelity ECG interpretation is increasingly reliant on massive foundation models, yet their deployment in clinical edge-care remains hindered by extreme computational demands. While knowledge distillation (KD) is a promising solution, traditional methods fail to capture the complex spatio-temporal dependencies of ECG signals when transferring knowledge across heterogeneous architectures. In this paper, we propose EVL-ECG, a framework specifically designed for cross-architecture distillation of cardiac diagnostic logic. EVL-ECG introduces three ECG-aware innovations: (1) Multi-Head Cross-Attention Alignment, which harmonizes architectural discrepancies to preserve fine-grained morphological features; (2) Optimal Transport-based Visual Feature Matching, utilizing optimal transport to maintain global structural relationships across ECG leads despite mismatched token representations; and (3) Geometric Intra-Architecture Relation Matching, which distills the latent diagnostic reasoning of the teacher model. Evaluations across ECG benchmarks demonstrate that EVL-ECG yields improvements of up to 2.4\% AUC and 1.1\% clinical accuracy over existing baselines. Notably, EVL-ECG establishes an efficient 2B-parameter ECG foundation model, suitable for resource-constrained clinical environments.
\end{abstract}

\section{Introduction}
\label{sec:intro}

Vision-Language Models (VLMs) have advanced Electrocardiogram (ECG) interpretation from simple classification to generative clinical reporting~\cite{ecg-survey,intro-ecg,ecg-gpt,pulse,gem,ecgchat}. However, their massive architectures prohibit real-time clinical deployment. While Knowledge Distillation (KD) offers a compression solution~\cite{minilm,llavakd,alignkd}, distilling from a VLM teacher to a smaller student faces two critical barriers: severe vocabulary mismatch due to tokenizer heterogeneity~\cite{minillm,emkd}, and unbalanced visual tokens caused by differing visual encoders.

Existing KD methods treat cross-tokenizer and visual-token mismatches in isolation~\cite{emo,mined,uld,cui2024multilevelot,emkd,alignkd,llavakd}, restricting the use of modern, efficient small language models as backbones. To bridge this gap, we propose EVL-ECG, a Multi-Aspect Heterogeneous KD framework that employs Multi-Head Cross-Attention to adaptively aggregate dense teacher representations into a cohesive feature space. This architecture utilizes Optimal Transport regularization to maintain vital global spatial and morphological ECG characteristics through soft visual token alignment, while simultaneously applying Geometric Intra-Architecture Relation Matching to distill complex diagnostic logic via distance- and angle-wise potentials, ensuring the student model masters the intricate clinical patterns inherent in ECG interpretation. We also discuss the related works in Appendix~\ref{sec:rel_works}. Our main contributions are:

\begin{enumerate}
\item We propose EVL-ECG, a unified heterogeneous knowledge distillation framework that resolves tokenizer and visual-token mismatches between massive teachers and efficient students through a Multi-Head Cross-Attention alignment and Optimal Transport regularization, ensuring the preservation of critical morphological features in ECG images.
\item We introduce a Geometric Intra-Architecture Relation Matching module that distills the teacher's internal diagnostic logic via distance- and angle-wise potentials, enabling the student model to inherit sophisticated reasoning patterns for complex cardiac analysis while maintaining a resource-friendly size.
\item Extensive evaluations across multiple ECG interpretation benchmarks demonstrate that EVL-ECG significantly outperforms state-of-the-art distillation methods and proprietary VLMs highlighting its robust clinical fidelity and strong generalizability to diverse diagnostic scenarios.
\end{enumerate}

\section{Method}
\begin{figure*}[t]
  \includegraphics[width=1.0\linewidth]{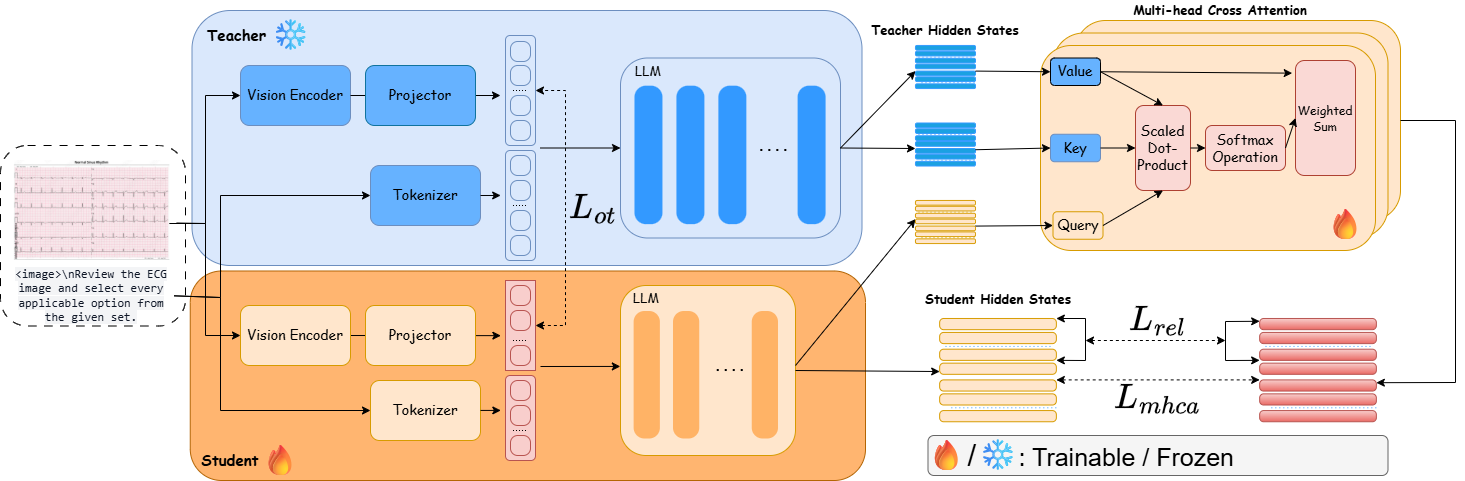} 
  \caption {Overview of the proposed \textbf{EVL-ECG} distillation framework. Our approach is tailored to capture the complex temporal and spatial dependencies of ECG signals. The framework employs multi-head cross-attention to align heterogeneous ECG representations between a large-scale teacher and an efficient student. Furthermore, it integrates OT-based visual matching to preserve the global structural patterns of ECG leads, while Geometric Intra-Architecture Relation Matching distills the underlying diagnostic logic essential for cardiac arrhythmia detection.}
  \label{fig:architecture}
\end{figure*}

This section details our proposed EVL-ECG framework in Figure~\ref{fig:architecture}, which facilitates multi-level heterogeneous knowledge distillation between a large teacher and an efficient student VLM.

\subsection{Optimal Transport-based Visual Feature Matching}\label{subsec:ot_image_alignment}In ECG images, spatial layout intrinsically encodes clinical information such as the standard 12-lead grid where specific patches correspond to distinct cardiac axes. Standard point-wise matching ignores this global spatial topology. To preserve the teacher's collective geometry of visual descriptors, ensuring the student accurately maps the spatial progression of leads and waveforms without positional confusion, we employ OT regularization. 

Let $\mathbf{t}_i, \mathbf{s}_j \in \mathbb{R}^{D_s}$ denote the $L_2$-normalized visual tokens of the teacher (linearly projected to the student's dimension) and student, respectively. We treat the $V_t$ teacher and $V_s$ student tokens as uniform empirical distributions: $\mu = \frac{1}{V_t} \sum_{i} \delta_{\mathbf{t}_i}$ and $\nu = \frac{1}{V_s} \sum_{j} \delta_{\mathbf{s}_j}$. Defining the transport cost as the squared Euclidean distance $C_{ij} = \| \mathbf{t}_i - \mathbf{s}_j \|_2^2$, we solve for the optimal transport plan $\mathbf{P}^\ast_\varepsilon \in \mathbb{R}_+^{V_t \times V_s}$ that minimizes the entropically regularized Sinkhorn distance:$$\mathbf{P}^\ast_\varepsilon = \arg\min_{\mathbf{P} \in \Pi(\mu, \nu)} \langle \mathbf{P}, C \rangle - \varepsilon \mathcal{H}(\mathbf{P})$$where $\Pi(\mu, \nu)$ is the set of valid couplings with marginals $\mu$ and $\nu$, and $\mathcal{H}(\mathbf{P})$ is the entropy of the transport plan. The OT-based distillation loss is then computed as the expected cost under this optimal coupling:$$\mathcal{L}_{ot} = \sum_{i=1}^{V_t} \sum_{j=1}^{V_s} P^\ast_{\varepsilon, ij} \| \mathbf{t}_i - \mathbf{s}_j \|_2^2$$By minimizing $\mathcal{L}_{ot}$, the student aligns its visual feature map with the teacher's distribution, actively preventing the spatial misalignment of critical diagnostic regions such as misinterpreting precordial leads as limb leads.

\subsection{Multi-Head Cross-Attention Alignment}

To bridge the sequence length disparity between the teacher's dense representations $H_t \in \mathbb{R}^{B \times L_t \times D_t}$, which capture high-resolution temporal ECG morphologies, and the student's compact latent space $H_s \in \mathbb{R}^{B \times L_s \times D_s}$, we employ Multi-Head Cross-Attention (MHCA). 

Treating the student’s hidden states as queries and the teacher’s states as keys and values allows the student to adaptively isolate and aggregate diagnostically relevant features, such as localized ectopic beats or subtle ST-segment deviations.

Rather than enforcing strict positional matching, the student dynamically attends to the most critical ECG segments across the sequence. The aligned teacher representation is obtained by computing $$\hat{H}_t = \text{Concat}(\text{head}_1, \dots, \text{head}_h)W^O$$ where the student features $H_s$ serve as the query and the teacher features $H_t$ act as both key and value, optimal $h$ detailed in Table~\ref{tab:hyper_param}. For each attention head, the operation is defined as $$\text{head}_i = \text{Softmax}\left(\frac{(H_s W_i^Q)(H_t W_i^K)^\top}{\sqrt{d_k}}\right)(H_t W_i^V)$$ where $W_i^Q$, $W_i^K$, and $W_i^V$ are learnable weight matrices. The final alignment loss ensures the student accurately reconstructs these clinical features by minimizing the mean squared error between its states and the attention-weighted teacher context:$$\mathcal{L}_{\mathrm{mhca}} = \frac{1}{B \cdot L_s} \sum_{b=1}^{B} \sum_{i=1}^{L_s} \| H_{s}^{(b,i)} - \hat{H}_t^{(b,i)} \|^2_2$$
\paragraph{Theoretical Motivation.} We demonstrate that the attention mechanism mathematically can be interpreted as an Entropic Barycentric Projection under an OT framework. In ECG interpretation, enforcing a strict 1-to-1 alignment is ill-posed when sequences differ in length ($L_s \neq L_t$) or when clinically relevant waveforms such as P waves, QRS complexes, and ST segments are temporally shifted or vary across leads. Further theoretical details are provided in Appendix~\ref{sec:theory}.

\subsection{Geometric Intra-Architecture Relation Matching}ECG interpretation relies fundamentally on the structural topology and temporal relationships between distinct waveform segments such as P-wave to QRS complex intervals, or ST-segment orientations. While point-wise alignment aids token reconstruction, it overlooks this global diagnostic logic. Inspired by relational KD~\cite{rkd}, we introduce a Geometric Intra-Architecture Relation Matching module to ensure the student maintains the teacher’s internal geometric reasoning regarding the ECG signal's morphology. 

We define two relation potentials for hidden state sequences $H$: a distance-wise potential $\psi_{\mathrm{D}}$ mean-normalized pairwise Euclidean distance to capture magnitude and interval-based signal relationships, and an angle-wise potential $\psi_{\mathrm{A}}$ pairwise cosine similarity to capture the directional morphology and electrical axis of the ECG features. The distillation loss aligns these structural potentials between the projected teacher $\hat{H}_t$ and student $H_s$ across batch $B$ and sequence length $L_s$:
\begin{multline}
\mathcal{L}_{k} = \frac{1}{B \cdot L_s^2} \sum_{b=1}^{B} \sum_{i,j=1}^{L_s} \biggl\| \psi_{k}(\hat{H}_{t}^{(b,i)}, \hat{H}_{t}^{(b,j)}) \\ 
- \psi_{k}(H_{s}^{(b,i)}, H_{s}^{(b,j)}) \biggr\|^2
\end{multline} 
where $k \in \{\mathrm{D}, \mathrm{A}\}$ denotes the distance and angle metrics. The total relational loss, $\mathcal{L}_{\mathrm{rel}} = \frac{1}{2}(\mathcal{L}_{\mathrm{D}} + \mathcal{L}_{\mathrm{A}})$, acts as a structural regularizer. By matching this geometric manifold, the student learns the teacher's clinical clustering, which is vital for recognizing complex, multi-segment cardiac abnormalities. To further illustrate our findings, we include visualizations in Appendix~\ref{sec:visual}. We also provide clinical insights and limitations of our method in Appendix~\ref{sec:clinical_insights_limitations}.

\subsection{Total Distillation Objective}

The final objective function integrates the cross-architecture alignment and the geometric constraints. By combining the MHCA-based reconstruction with relation matching, the student model effectively inherits both the fine-grained diagnostic features and the global reasoning patterns of the teacher model:
$$
\mathcal{L}_{total} = (1 - \alpha)\mathcal{L}_{CE} + \alpha\cdot( \lambda_{m} \mathcal{L}_{mhca} + \lambda_{r} \mathcal{L}_{rel} + \lambda_{ot} \mathcal{L}_{ot})
$$
 Best hyper-parameters after the searching process are provided in Table~\ref{tab:hyper_param} from Appendix~\ref{sec:appendix:impl}.

\begin{table*}[h]
\centering
\caption{Comparison of EVL-ECG with state-of-the-art domain-specific methods, proprietary MLLMs, and open-source MLLMs across ECG benchmarks. $\downarrow$ indicates that lower values correspond to superior model performance.}
\label{tab:main_results}
{\fontsize{8}{10}\selectfont
\setlength{\tabcolsep}{3.5pt}
\begin{tabular}{l ccc ccc c ccc c c c}
\toprule
\textbf{Datasets} &
\multicolumn{3}{c}{\textbf{PTB-XL-Super}} &
\multicolumn{3}{c}{\textbf{CODE-15\%}} &
\textbf{ECG-QA} &
\multicolumn{3}{c}{\textbf{CPSC-2018}} &
\textbf{CSN} &
\textbf{G12EC} &
\textbf{MMMU-ECG} \\
\midrule
\textbf{Metric} &
AUC & F1 & HL $\downarrow$ &
AUC & F1 & HL $\downarrow$ &
Accuracy &
AUC & F1 & HL $\downarrow$ &
Accuracy &
Accuracy &
Accuracy \\
\midrule
Random &
50.3 & 33.2 & 50.1 &
48.8 & 15.0 & 32.1 &
16.2 &
51.2 & 15.1 & 28.8 &
11.6 &
12.1 &
24.2 \\
\midrule
\rowcolor[gray]{0.9} \multicolumn{14}{l}{\textit{Proprietary VLMs}} \\
GPT-4o &
55.6 & 28.3 & 26.2 &
59.9 & 24.9 & 15.7 &
35.2 &
50.9 & 10.6 & 18.2 &
57.5 &
49.2 &
43.5 \\
Gemini 1.5 Pro &
50.7 & 15.3 & 27.9 &
56.7 & 20.0 & 15.9 &
33.2 &
50.1 & 7.4 & 20.5 &
50.5 &
36.0 &
40.0 \\
Claude 3.5 Sonnet &
54.0 & 27.5 & 29.6 &
58.3 & 20.3 & 17.8 &
34.2 &
52.8 & 11.5 & 18.9 &
51.5 &
51.4 &
42.0 \\
\midrule
\rowcolor[gray]{0.9} \multicolumn{14}{l}{\textit{Open-source VLMs}} \\
LLaVA-Med &
50.0 & 12.3 & 28.1 &
69.2 & 27.0 & 33.4 &
29.5 &
50.0 & 2.5 & 20.2 &
13.8 &
14.1 &
27.0 \\
LLaVA-OneVision-7B &
49.8 & 11.4 & 34.5 &
58.7 & 17.0 & 20.6 &
20.4 &
49.6 & 8.0 & 28.3 &
23.3 &
25.7 &
26.0 \\
LLaVA-1.6-Vicuna-13B &
50.0 & 20.1 & 38.3 &
53.0 & 3.6 & 16.6 &
22.0 &
50.0 & 19.3 & 62.8 &
31.4 &
35.0 &
38.0 \\
LLaVA-1.6-34B &
50.2 & 19.9 & 36.0 &
57.2 & 12.8 & 16.6 &
22.4 &
49.6 & 19.3 & 62.8 &
44.3 &
45.9 &
31.0 \\
\midrule
\rowcolor[gray]{0.9} \multicolumn{14}{l}{\textit{Domain-specific VLMs/KD Methods}} \\
ECG-GPT &
69.5 & 53.9 & 20.1 &
68.9 & 40.1 & 17.4 &
N/A &
\textbf{69.3} & \textbf{44.0} & \textbf{9.9} &
N/A &
N/A &
N/A \\
SFT &
\underline{74.6} & 62.8 & \underline{16.3} &
85.0 & 76.7 & 7.0 &
63.7 &
66.4 & 32.7 & 10.8 &
\underline{89.3} &
\underline{76.4} &
\underline{47.7} \\
ULD~\cite{uld} &
73.4 & 61.7 & 16.9 &
84.3 & 76.4 & 7.2 &
62.5 &
65.5 & 28.2 & 11.6 &
86.3 &
75.6 &
46.5 \\
MultiLevelOT~\cite{cui2024multilevelot} &
72.3 & 59.4 & 17.1 &
\underline{85.6} & 74.7 & 7.4 &
63.1 &
66.6 & 29.1 & 12.1 &
87.9 &
76.1 &
44.5 \\
DSKD~\cite{dskd} &
73.0 & 60.3 & 18.8 &
80.8 & 75.4 & 8.0 &
60.5 &
65.3 & 31.3 & 12.1 &
88.4 &
75.8 &
46.5 \\
EM-KD~\cite{emkd} &
74.3 & \underline{62.9} & 16.4 &
84.9 & \underline{77.4} & \underline{6.8} &
\underline{63.9} &
66.3 & 32.9 & 11.2 &
89.2 &
76.0 &
47.4 \\
\textbf{EVL-ECG (Ours)} &
\textbf{75.2} & \textbf{63.3} & \textbf{16.0} &
\textbf{86.4} & \textbf{78.4} & \textbf{6.5} &
\textbf{64.8} &
\underline{66.8} & \underline{33.6} & \underline{12.3} &
\textbf{89.7} &
\textbf{77.1} &
\textbf{48.5} \\
\bottomrule
\end{tabular}
}
\end{table*}

\section{Experiments}

\subsection{Datasets and Metrics}

The training of EVL-ECG leverages a large-scale dataset, which is ECGInstruct from PULSE~\cite{pulse} with 1,156,110 conversations. This data comprises four primary sources: PTB-XL~\cite{ptb-xl}, ECG-QA~\cite{ecg-qa}, MIMIC-IV-ECG~\cite{mimic-iv}, and CODE-15\%~\cite{code15}.

To evaluate the clinical diagnostic capabilities of EVL-ECG, we utilize ECG-Bench~\cite{pulse}, a standardized dataset designed for the assessment of VLMs in the cardiac domain, which compresses from multiple sources. Following PULSE~\cite{pulse}, we utilize multi-label classification metrics, including Macro AUC, Macro F1, and Hamming Loss, to evaluate the datasets PTB-XL Super, CODE-15\%, and CPSC-2018~\cite{cpsc-g12ec}, where multiple correct labels may exist. For the ECG-QA, CSN~\cite{csn}, MMMU-ECG~\cite{pulse} and G12EC~\cite{cpsc-g12ec} datasets, we adopt accuracy as the evaluation metric. Evaluation datasets statistics are reported in Appendix~\ref{sec:appendix:data_statistic}.

\subsection{Baselines}

We compare our method against established methods including other knowledge distillation methods, domain-specific methods and MLLMs with different size. We consider three proprietary MLLMs to establish a high-level performance benchmark: GPT-4o~\cite{gpt4o}, Gemini 1.5 Pro~\cite{gemini1.5}, and Claude 3.5 Sonnet~\cite{sonnet}. For open-source VLMs, we select a range of open-source models to ensure comprehensive coverage across different visual components. These include the general-purpose LLaVA-1.6~\cite{llavanext}, LLaVA-OneVision-7B~\cite{llava-ov}, and the domain-specific LLaVA-Med~\cite{llava-med}. With domain-specific methods and KD baselines, we evaluate our model against ECG-GPT~\cite{ecg-gpt}, supervised fine-tuning (SFT) and other state-of-the-art KD frameworks, which are ULD~\cite{uld}, MultiLevel-OT~\cite{cui2024multilevelot}, DSKD~\cite{dskd} and EM-KD~\cite{emkd}. More implementation details are reported in the Appendix~\ref{sec:appendix:impl} due to page limitation.

\subsection{Main Results}

EVL-ECG demonstrates a clear performance advantage over both high-tier proprietary models and open-source VLMs across benchmarks, as detailed in Table \ref{tab:main_results}. While general-purpose models struggle with intricate temporal ECG patterns, evidenced by EVL-ECG's 75.2 AUC on PTB-XL-Super compared to GPT-4o’s 55.6, our model also introduces a unified knowledge distillation framework that simultaneously resolves tokenizer and visual token mismatches. This framework allows EVL-ECG to consistently outperform established KD baselines, while maintaining greater cross-dataset versatility than specialized models like ECG-GPT. Beyond aggregate metrics, we observe that EVL-ECG yields more consistent predictions across correlated leads, indicating improved modeling of inter-lead dependencies that are critical for clinical diagnosis. In particular, the model shows enhanced sensitivity to subtle morphological variations, such as ST-segment deviations and irregular QRS complexes, which are commonly associated with clinically significant cardiac abnormalities. We also report additional experiments and loss components contributions in table~\ref{tab:ablation_study} from Appendix~\ref{sec:analysis} respectively.



\section{Conclusion and Future Works}

This work presents EVL-ECG, an efficient Vision-Language Model achieving state-of-the-art performance in ECG interpretation through a specialized hierarchical distillation framework. By bridging heterogeneous architectures via the MHCA module and ensuring structural integrity with OT-based and geometric matching, our model captures the complex diagnostic logic required for high-fidelity clinical accuracy. This enables the deployment of sophisticated reasoning patterns in resource-constrained medical environments. Future research will focus on enhancing model robustness against real-world clinical noise, such as baseline wander and electrode artifacts, to ensure reliability in ambulatory settings. Additionally, we aim to extend this distillation paradigm to capture multi-lead temporal dependencies, further aligning distilled representations with the nuanced patterns used in expert cardiological diagnosis. 

\section*{Acknowledgment}
This work was funded by Vingroup Joint Stock Company (Vingroup JSC), Vingroup, and supported by Vingroup Innovation Foundation (VINIF) under project code VINIF.2021.DA00128.

\nocite{langley00}

\bibliography{example_paper}
\bibliographystyle{icml2026}

\newpage
\appendix
\onecolumn
\section{Related Works}
\label{sec:rel_works}
\subsection{Multimodal Large Language Models}
The rapid evolution of MLLMs, such as the Qwen-VL series~\cite{Qwen3-VL} and LLaVA-OneVision~\cite{llava-ov}, has expanded LLMs' sensory scope by using learnable projectors to translate visual features into linguistic workspaces. This multimodality is particularly critical in healthcare~\cite{surveymedllm}, where clinical decision-making inherently relies on integrating multidimensional data like medical imaging, physiological time-series, and narrative reports~\cite{ecg-lm}.

\subsection{Knowledge Distillation}
Knowledge distillation (KD) compresses models by transferring capabilities from a teacher to a student~\cite{origkd}. For VLMs, KD strategies primarily divide into LLM-style and MLLM-style. LLM-style approaches refine text generation and reduce hallucinations~\cite{minillm}, with recent advances utilizing Optimal Transport for cross-tokenizer and sequence alignment~\cite{uld, cui2024multilevelot, mined, dskd}. Conversely, MLLM-style methods prioritize visual feature and cross-modal consistency~\cite{llavakd, alignkd}. Notably, frameworks like EM-KD~\cite{emkd} address general distillation scenarios with unbalanced vision tokens.

\subsection{ECG Analysis}
AI-driven ECG interpretation has shifted from retrieval-augmented zero-shot models~\cite{yu-ecg} toward complex multimodal alignment. Frameworks like JoLT~\cite{cai2023jolt} and ECG-CoCa~\cite{ecgchat} synchronize 1D ECG signals with text, whereas PULSE~\cite{pulse} synthesizes ECG images to leverage VLM architectures. Recent developments further incorporate instruction-tuning for clinical report generation~\cite{meit} and unified architectures like GEM~\cite{gem}, which process temporal signals and visual waveforms simultaneously to mimic a cardiologist's workflow.

\section{Theoretical Motivation of MHCA}
\label{sec:theory}
\textit{Let the student queries $Q_m^{(i)}$ and teacher keys $K_n^{(i)}$ represent empirical samples from latent feature distributions. For a given head $i$, define the transport cost between the $m$-th student token and $n$-th teacher token as the negative inner product $C_{mn}^{(i)} = - Q_m^{(i)} (K_n^{(i)})^\top$. Then, the cross-attention mechanism computes the optimal conditional transport plan $\pi^*$ that minimizes the Entropic Optimal Transport objective, and the resulting context vector $\text{head}_m^{(i)}$ is the exact Barycentric Projection of the teacher's value space onto the student's coordinate system.}

Consider the $m$-th student token. We seek a probability distribution or conditional transport plan $\pi_m \in \Delta^{L_t-1}$ over the teacher tokens $n \in \{1, \dots, L_t\}$ that minimizes the expected transport cost from the student to the teacher, while maintaining a degree of smoothness. We formulate this as an entropy-regularized optimization problem:
$
\pi_m^* = \arg\min_{\pi_m \in \Delta^{L_t-1}} \left( \sum_{n=1}^{L_t} \pi_{mn} C_{mn}^{(i)} - \epsilon \mathcal{H}(\pi_m) \right)
$
where $\Delta^{L_t-1}$ is the $(L_t-1)$-dimensional probability simplex, $C_{mn}^{(i)} = - Q_m^{(i)} (K_n^{(i)})^\top$ is the geometric cost function, $\mathcal{H}(\pi_m) = -\sum_n \pi_{mn} \log \pi_{mn}$ is the Shannon entropy acting as a regularizer, and $\epsilon > 0$ is the regularization coefficient. To solve this constrained optimization problem, we introduce a Lagrange multiplier $\lambda$ for the simplex constraint $\sum_{n=1}^{L_t} \pi_{mn} = 1$:
\begin{equation}
\mathcal{L}(\pi_m, \lambda) = \sum_{n=1}^{L_t} \pi_{mn} C_{mn}^{(i)} + \epsilon \sum_{n=1}^{L_t} \pi_{mn} \log \pi_{mn} + \lambda \left( \sum_{n=1}^{L_t} \pi_{mn} - 1 \right)
\end{equation}
Taking the partial derivative with respect to $\pi_{mn}$ and setting it to zero yields:
\begin{equation}
\frac{\partial \mathcal{L}}{\partial \pi_{mn}} = C_{mn}^{(i)} + \epsilon (1 + \log \pi_{mn}) + \lambda = 0
\end{equation}
Solving for $\pi_{mn}$, we obtain
$
\pi_{mn}^* = \exp\left( -\frac{C_{mn}^{(i)}}{\epsilon} - \frac{\lambda}{\epsilon} - 1 \right) \propto \exp\left( -\frac{C_{mn}^{(i)}}{\epsilon} \right)
$ and applying the sum-to-one constraint $\sum_{j} \pi_{mj}^* = 1$ to find the normalizing constant gives:
$$
\pi_{mn}^* = \frac{\exp\left( - C_{mn}^{(i)} / \epsilon \right)}{\sum_{j=1}^{L_t} \exp\left( - C_{mj}^{(i)} / \epsilon \right)}
$$
By substituting our defined cost $C_{mn}^{(i)} = - Q_m^{(i)} (K_n^{(i)})^\top$ and setting the entropy regularization coefficient to $\epsilon = \sqrt{d_k}$, we exactly recover the attention weights from the original formulation:
$$
\pi_{mn}^* = \frac{\exp\left( \frac{Q_m^{(i)} (K_n^{(i)})^\top}{\sqrt{d_k}} \right)}{\sum_{j=1}^{L_t} \exp\left( \frac{Q_m^{(i)} (K_j^{(i)})^\top}{\sqrt{d_k}} \right)} \equiv w_{mn}^{(i)}
$$

In optimal transport, once the optimal plan $\pi^*$ is found, mapping the source distribution (teacher) to the target support (student) in a metric space is achieved via the barycentric projection. The projection of the teacher's features as values $V^{(i)}$ onto the $m$-th student token is defined as the expectation under the optimal transport plan:
$$
\mathcal{P}_{\pi^*}(V^{(i)})_m = \mathbb{E}_{n \sim \pi_m^*} \left[ V_n^{(i)} \right] = \sum_{n=1}^{L_t} \pi_{mn}^* V_n^{(i)} = \sum_{n=1}^{L_t} w_{mn}^{(i)} V_n^{(i)} = \text{head}_m^{(i)}
$$
Thus, our final loss function $\mathcal{L}_{mhca}$ is mathematically equivalent to minimizing the $L_2$ distance between the student's representation and the optimal entropic barycentric projection of the teacher's representation.

\section{Experiment Details}
\subsection{Implementation Details}
\label{sec:appendix:impl}
We employ \textbf{Qwen3-VL-2B-Instruct}~\cite{Qwen3-VL} as the student and \textbf{PULSE-7B}~\cite{pulse} as the teacher (details in Tables~\ref{tab:training_configs_sft} and~\ref{tab:training_configs_kd}). Training involves two phases: (1) \textbf{Supervised fine-tuning} for 54,195 steps with batch size 64 for domain adaptation, establishing a baseline checkpoint; and (2) \textbf{Knowledge Distillation} applied to this initialization. All baseline KD methods are trained in 1 epoch with \textbf{three different runs} and reported average score, start from this same checkpoint for fair comparison. We use \textbf{full-model fine-tuning} across both phases to ensure the student thoroughly learns complex ECG diagnostic patterns. Models are implemented in \textbf{PyTorch} (Python 3.11) and trained on a single \textbf{NVIDIA H100 (80GB)} using \texttt{bf16} mixed-precision. 

\vspace{0.2cm}

\begin{table}[h]
\centering
\caption{Training configurations for the SFT phase.}
\label{tab:training_configs_sft}
\begin{tabular}{lc}
\toprule
\textbf{Settings} & \textbf{Qwen3-VL-2B-Instruct} \\
\midrule
Steps            & 54195                \\
Learning Rate    & $2\times10^{-4}$ \\
Projector LR     & $5\times10^{-4}$ \\
Global Batch Size       & 64               \\
BF16 Setting & True \\
LR Scheduler     & Cosine           \\
Warmup Ratio     & 0.03             \\
Weight Decay     & 0.01             \\
Model Max Length        & 4096               \\
Min Pixels/Max Pixels & 784/50176              \\
\bottomrule
\end{tabular}
\end{table}

\begin{table}[h]
\centering
\caption{Training configurations for the knowledge distillation phase, also for all the KD baseline methods.}
\label{tab:training_configs_kd}
\begin{tabular}{lc}
\toprule
\textbf{Settings} & \textbf{Qwen3-VL-2B-Instruct} \\
\midrule
Epoch            & 1                \\
Learning Rate    & $2\times10^{-4}$ \\
Projector LR     & $5\times10^{-4}$ \\
Global Batch Size       & 48               \\
BF16 Setting & True \\
LR Scheduler     & Cosine           \\
Warmup Ratio     & 0.03             \\
Weight Decay     & 0.01             \\
Model Max Length        & 4096               \\
Min Pixels/Max Pixels & 784/50176              \\
\bottomrule
\end{tabular}
\end{table}

\paragraph{Hyperparameters} We fixed the Sinkhorn regularization ($\lambda$) parameter to 0.1, a widely adopted standard value in Optimal Transport implementations to ensure numerical stability. The weight hyper-parameters are 
selected from the following ranges: $\alpha  \in [0.1, 0.3, 0.5, 0.7, 1.0]$, $\lambda_m  \in [0.1, 0.3, 0.5, 0.7, 1.0]$, $\lambda_r \in [0.01, 0.03, 0.05, 0.07]$, $\lambda_{ot} \in [0.1, 0.3, 0.5]$ and $h \in [4, 8, 16]$.

\begin{table}[h]
\centering
\vspace{2mm}
\small
\setlength{\tabcolsep}{15pt} 
\caption{Optimal hyperparameters for our proposed framework.}
\begin{tabular}{lccccc}

\toprule
\textbf{Hyperparameters} & $\mathbf{\alpha}$ & $\mathbf{\lambda_\textit{m}}$ & $\mathbf{\lambda_\textit{r}}$ & $\mathbf{\lambda_\textit{ot}}$ &
$h$\\ \midrule
Best Values        & 0.3         & 1.0                & 0.03         & 0.1   & 8               \\
\bottomrule
\end{tabular}
\label{tab:hyper_param}
\end{table}

\paragraph{Knowledge Distillation Baselines} To the best of our knowledge, our work is the first unified framework to simultaneously address the dual alignment challenge: the discrepancy in both text tokenizers and visual token counts in VLMs. Consequently, we select and adapt state-of-the-art baselines from both cross-tokenizer LLM distillation and cross-visual token VLM distillation to compare with our framework, which are DSKD~\cite{dskd},  ULD~\cite{uld}, MultiLevelOT~\cite{cui2024multilevelot} and EM-KD~\cite{emkd}.

\begin{figure*}[t]
  \centering
  \includegraphics[width=0.6\linewidth]{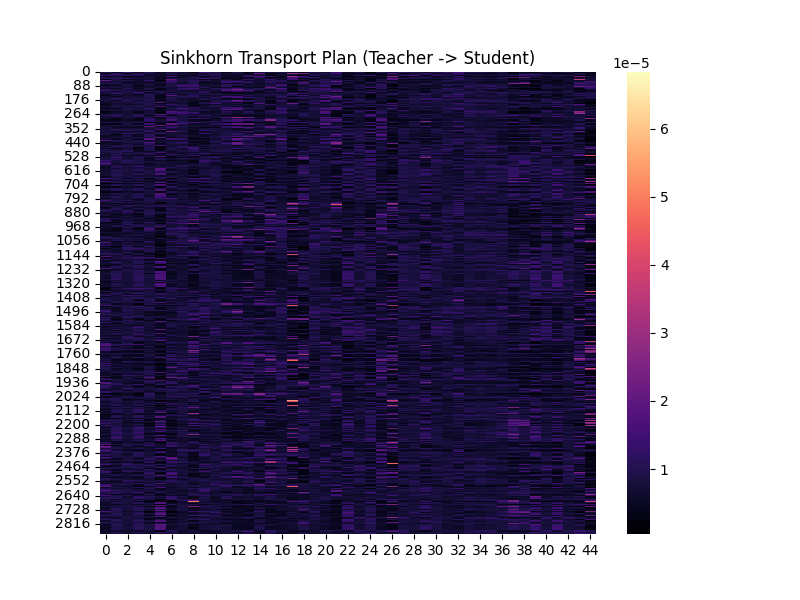} 
    \caption{Heatmap showing the Sinkhorn Transport Plan between teacher and student features. The intensity of each cell represents the transport cost between the corresponding feature representations, with brighter colors indicating higher similarity and lower costs.}
    \label{fig:ot_analysis}
\end{figure*}

\subsection{Datasets}
\label{sec:appendix:data_statistic}

\begin{table}[htbp]
\centering
\caption{Overview of evaluation datasets in ECGBench. This collection contains both in-domain and out-of-domain problems across tasks with diverse answer types.}
\label{tab:evaluation_datasets}
\small 
\setlength{\tabcolsep}{5pt}
\begin{tabular}{lllr c} 
\toprule
\textbf{Evaluation Dataset} & \textbf{Task} & \textbf{Type} & \textbf{\# Samples} & \textbf{In-Domain?} \\ 
\midrule
PTB-XL Super & Abnormality Detection & Close-ended & 2,082 & YES \\
CODE-15\% & Abnormality Detection & Close-ended & 1,400 & YES \\
ECG-QA & Abnormality Detection & Close-ended & 1,317 & YES \\ 
\midrule
CPSC 2018 & Abnormality Detection & Close-ended & 2,061 & NO \\
CSN & Abnormality Detection & MCQ (8-option) & 1,611 & NO \\
G12EC & Abnormality Detection & MCQ (8-option) & 2,026 & NO \\
MMMU ECG & Multimodal Understanding & MCQ (4-option) & 200 & NO \\
\bottomrule
\end{tabular}
\end{table}

For the training dataset, we utilize ECGInstruct as we mentioned in the previous sections. To evaluate the clinical diagnostic capabilities of EVL-ECG, we utilize ECGBench, which contains both repurposed tasks from different existing
datasets. Table~\ref{tab:evaluation_datasets} shows the details of each evaluation dataset.

\section{Visualizations}
\label{sec:visual}


Figure~\ref{fig:ot_analysis} visualizes the Sinkhorn Transport Plan for visual feature alignment. By minimizing this transport cost, the student model is forced to accurately replicate the teacher's precise spatial understanding of ECG topology. This ensures that critical clinical features, such as specific lead placements and localized PQRST waveforms, are correctly mapped without spatial misalignment, facilitating robust diagnostic knowledge transfer.

\begin{figure*}[t]
  \centering
  \includegraphics[width=0.6\linewidth]{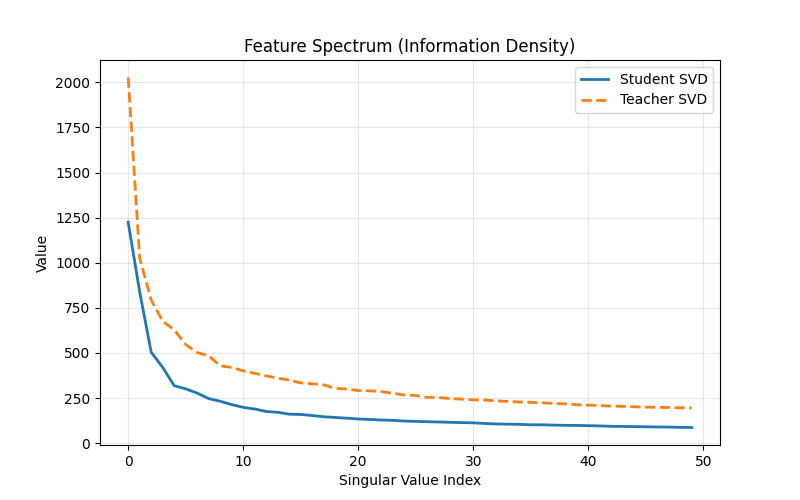} 
    \caption{Comparison of the Feature Spectrum via Singular Value Decomposition (SVD). The plot depicts the singular value decay for both Student and Teacher models, reflecting the information density and rank distribution of their respective feature representations.}
    \label{fig:svd}
\end{figure*}

Singular Value Decomposition of the feature matrices reveals a rapid decay for both models, indicating that core ECG diagnostic patterns which are primary waveforms and cardiac axes occupy a low-rank subspace. While the teacher's higher singular values reflect a richer capture of subtle clinical nuances, the structurally similar decay curves confirm the student effectively learns to prioritize the same critical cardiac features.

\section{Analysis} \label{sec:analysis}

\begin{table}[h]
\centering
\small
\caption{Ablation study of proposed loss components across different ECG benchmarks.}
\label{tab:ablation_study}
\begin{tabular}{l ccc ccc cc@{}}
\toprule
\textbf{Method} & $\mathcal{L}_{ot}$ & $\mathcal{L}_{mhca}$ & $\mathcal{L}_{rel}$ & \multicolumn{3}{c}{\textbf{CODE-15\%}} & \textbf{ECG-QA} & \textbf{G12EC} \\
\midrule
& & & & AUC & F1 & HL & Acc & Acc \\
\midrule
Baseline & & & & 85.0 & 76.7 & 7.0 & 63.7 & 76.4 \\
\midrule
\multirow{3}{*}{Components} & \checkmark & & & 85.4 & 77.1 & 7.1 & 64.1 & 76.7 \\
& \checkmark & \checkmark & & 85.9 & 77.8 & 6.8 & 64.5 & 76.8 \\
& \checkmark & \checkmark & \checkmark & \textbf{86.4} & \textbf{78.4} & \textbf{6.5} & \textbf{64.8} & \textbf{77.1} \\
\bottomrule
\end{tabular}
\end{table}

\paragraph{Impact of Loss Components.} 

The ablation study presented in Table \ref{tab:ablation_study} quantifies the incremental performance gains provided by each proposed loss component across three distinct ECG benchmarks. Starting from a baseline of 85.0\% AUC on CODE-15\% and 63.7\% accuracy on ECG-QA, the introduction of the OT loss ($\mathcal{L}_{ot}$) establishes a foundation for geometric alignment, marginally improving most metrics. A more substantial performance leap occurs with the integration of the MHCA reconstruction loss ($\mathcal{L}_{mhca}$), which elevates the CODE-15\% AUC to 85.9 and G12EC accuracy to 76.8\%, underscoring its efficacy in aligning cross-architecture feature representations. The full configuration, which incorporates the relational matching loss ($\mathcal{L}_{rel}$), achieves the superior result in every category, reaching peak values of 86.4 AUC, 78.4 F1, and a record-low 6.5 Hamming Loss on CODE-15\%, while also maximizing accuracy on ECG-QA and G12EC. 

\paragraph{Computational Analysis.}
We evaluate the computational efficiency of the proposed EVL-ECG method by comparing its per-step training time and peak GPU memory (VRAM) usage against different KD method baselines. As summarized in Table~\ref{tab:training_time}, EVL-ECG maintains a competitive computational footprint, requiring only 8.54 seconds per step and 56.4 GB of VRAM. While slightly higher than the baseline ULD (8.37s, 55.7 GB), EVL-ECG remains significantly more efficient than EM-KD, which exhibits the highest overhead at 9.13s and 77.4 GB. These results demonstrate that EVL-ECG provides an effective trade-off, achieving its performance gains without necessitating the prohibitive hardware requirements or extended training durations seen in more complex distillation frameworks.

\begin{table}[h]
\centering
\vspace{2mm}
\small
\setlength{\tabcolsep}{4pt} 
\caption{Computation time and GPU memory consumption of different KD methods.}
\begin{tabular}{lccccc}

\toprule
\textbf{Method} & \textbf{ULD} & \textbf{MultiLevelOT} & \textbf{DSKD} & \textbf{EM-KD} &  \textbf{EVL-ECG} \\ \midrule
Time/step (s)        & 8.37         & 8.64                & 8.48         & 9.13           & 8.54         \\
VRAM (GB)       & 55.7         & 58.1                & 57.4         & 77.4           & 56.4         \\ \bottomrule
\end{tabular}
\label{tab:training_time}
\end{table}


\section{Clinical Insights and Limitations}
\label{sec:clinical_insights_limitations}

\paragraph{Clinical Insights.}
EVL-ECG is designed to preserve clinically meaningful structure in ECG interpretation rather than merely improve aggregate prediction scores. In particular, the combination of cross-attention alignment, optimal transport matching, and relational distillation helps the student model retain inter-lead consistency and waveform geometry, which are essential for recognizing patterns such as ST-segment deviation, QRS morphology changes, and rhythm irregularities. This is especially valuable in ECG analysis because many clinically important abnormalities are expressed through subtle temporal shifts or lead-wise correlations that can be missed by compact models trained with point-wise supervision alone.

\paragraph{Clinical Limitations.}
Despite these advantages, EVL-ECG remains a decision-support model rather than a clinical diagnostic system. Its performance is evaluated on curated benchmarks, and real-world ECGs may contain noisy signals, missing leads, acquisition artifacts, or patient-specific confounders that can reduce reliability. In addition, while the model improves structural alignment, it does not provide explicit causal explanations or guaranteed uncertainty calibration for every prediction. Therefore, deployment in clinical settings should be accompanied by clinician oversight, external validation on hospital-specific data, and careful assessment under distribution shift.

\end{document}